\documentclass[10pt,twocolumn,letterpaper]{article}

\usepackage{cvpr}
\usepackage{times}
\usepackage{epsfig}
\usepackage{graphicx}
\usepackage{amsmath,amssymb, bm} %
\usepackage[position=top]{subfig}
\usepackage{floatrow}
\usepackage{color}
\usepackage[ruled]{algorithm2e}
\usepackage{multirow}
\usepackage{tikz}
\def\bal#1\eal{\begin{align}#1\end{align}} %

\newcommand{\pr}[1]{\left(#1\right)} %

\def\transp{\mathsf{T}} %
\def\m{\mathbf}

\def\R{\mathbb{R}}

\DeclareMathOperator*{\argmin}{arg\,min}
\newcommand{\grad}{\ensuremath{\nabla}} %
\newcommand{\norm}[2]{\ensuremath{\left\|#1\right\|_{#2}}}

\newcommand {\bbmtx}{\begin{bmatrix}} %
\newcommand {\ebmtx}{\end{bmatrix}} %

\DeclareMathOperator{\prox}{prox}

\usepackage[export]{adjustbox}

\usepackage[breaklinks=true,bookmarks=false]{hyperref}
\hypersetup{
  colorlinks   = true, %
  urlcolor     = blue, %
  linkcolor    = blue, %
  citecolor   = green %
}
\cvprfinalcopy %

\def\cvprPaperID{2823} %

\begin{document}

\title{Iterative Residual CNNs for Burst Photography Applications}

\author{Filippos Kokkinos \qquad Stamatios Lefkimmiatis\\
Skolkovo Institute of Science and Technology (Skoltech), Moscow, Russia\\
{\tt\small filippos.kokkinos@skoltech.ru \qquad s.lefkimmiatis@skoltech.ru}
}

\maketitle

\begin{figure*}[h]
  \centering
  \subfloat[Input]{\includegraphics[width=0.24\textwidth,valign=c]{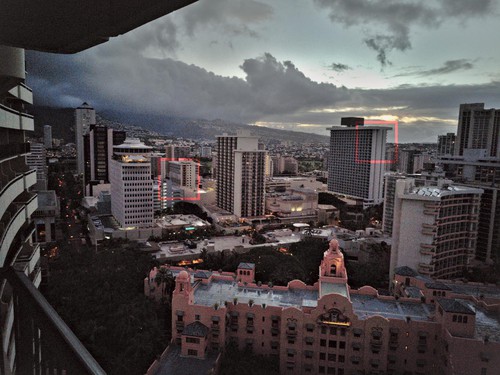}} \hspace{0.005\textwidth}
  \subfloat[Kokkinos~\cite{kokkinos2018iterative}]{\includegraphics[width=0.24\textwidth,valign=c]{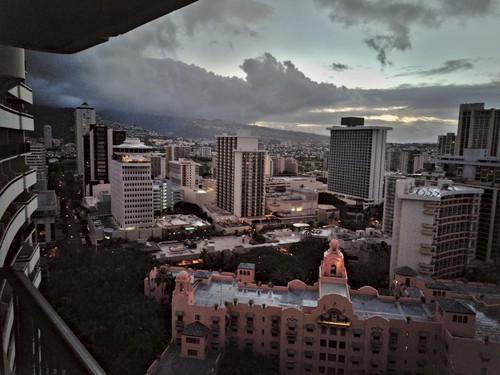}} \hspace{0.005\textwidth}
  \subfloat[HDR+~\cite{hasinoff2016burst}]{\includegraphics[width=0.24\textwidth,valign=c]{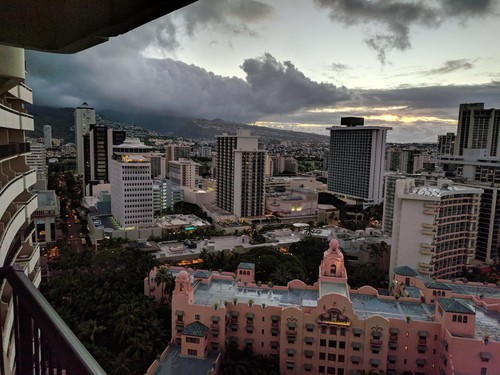}} \hspace{0.005\textwidth}
  \subfloat[Ours]{\includegraphics[width=0.24\textwidth,valign=c]{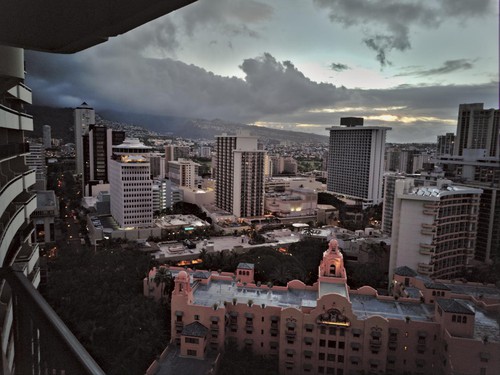}}
  \vspace{-0.5cm}
  \\
  \subfloat{\subfloat {\includegraphics[width=0.1169\textwidth,valign=c]{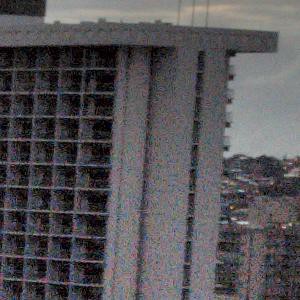}} \hspace{0.001\textwidth}
                          \subfloat {\includegraphics[width=0.1169\textwidth,valign=c]{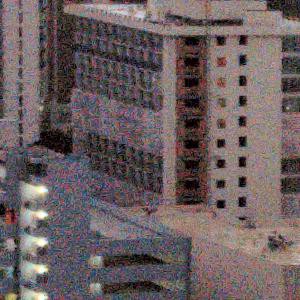}}} \hspace{0.005\textwidth}
  \subfloat{\subfloat {\includegraphics[width=0.1169\textwidth,valign=c]{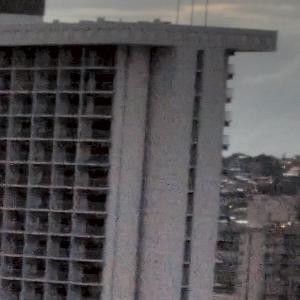}} \hspace{0.001\textwidth}
                          \subfloat {\includegraphics[width=0.1169\textwidth,valign=c]{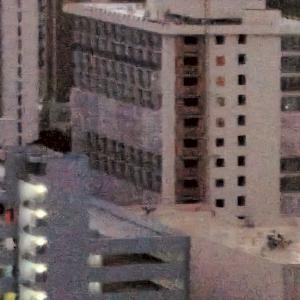}}} \hspace{0.005\textwidth}
  \subfloat{\subfloat {\includegraphics[width=0.1169\textwidth,valign=c]{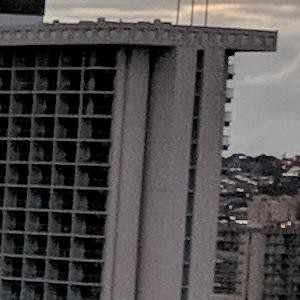}} \hspace{0.001\textwidth}
                          \subfloat {\includegraphics[width=0.1169\textwidth,valign=c]{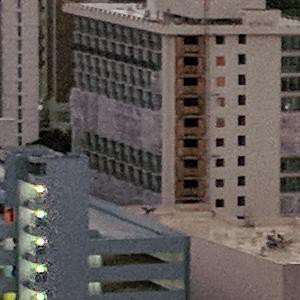}}} \hspace{0.005\textwidth}
  \subfloat{\subfloat{\includegraphics[width=0.1169\textwidth,valign=c]{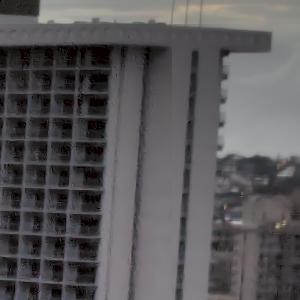}} \hspace{0.001\textwidth}
                          \subfloat {\includegraphics[width=0.1169\textwidth,valign=c]{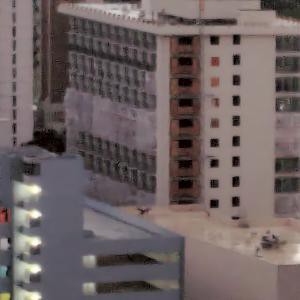}}}
  \vspace{-0.0cm}
  \caption{Demosaicking and denoising of a real low-light raw burst from the HDR+ dataset~\cite{hasinoff2016burst}. Our method achieves high quality reconstruction even in cases of excessive noise in the sensor data.} 
  \label{fig:real_image}
  \vspace{-.35cm}
\end{figure*}

\begin{abstract}
Modern inexpensive imaging sensors suffer from inherent hardware constraints which often result in captured images of poor quality. Among the most common ways to deal with such limitations is to rely on burst photography, which nowadays acts as the backbone of all modern smartphone imaging applications. In this work, we focus on the fact that every frame of a burst sequence can be accurately described by a forward (physical) model. This, in turn, allows us to restore a single image of higher quality from a sequence of low-quality images as the solution of an optimization problem. Inspired by an extension of the gradient descent method that can handle non-smooth functions, namely the proximal gradient descent, and modern deep learning techniques, we propose a convolutional iterative network with a transparent architecture. Our network uses a burst of low-quality image frames and is able to produce an output of higher image quality recovering fine details which are not distinguishable in any of the original burst frames. We focus both on the burst photography pipeline as a whole, i.e., burst demosaicking and denoising, as well as on the traditional Gaussian denoising task. The developed method demonstrates consistent state-of-the art performance across the two tasks and as opposed to other recent deep learning approaches does not have any inherent restrictions either to the number of frames or their ordering. 
\end{abstract}

\vspace{-.5cm}
\section{Introduction}
\label{sec:Introduction}
With more than one billion smartphones sold each year, smartphone cameras have dominated the photography market. However, to allow for small and versatile sensors, inevitably manufacturers of such cameras need to make several compromises. As a result, the quality of images captured by smartphone cameras is significantly inferior compared to the quality of images acquired by sophisticated hand-held cameras like DSLRs. The most common hardware restrictions in smartphone cameras are the lack of large aperture lens and the small sensors that consist of fewer photodiodes. To overcome such inherent hardware restrictions, the focus is thus shifted towards the software of the camera, i.e., the Image Processing Pipeline (ISP). 

The shortcomings of mobile photography can be mitigated by the use of burst photography, where a camera firstly captures a burst of images, milliseconds apart, and afterward fuses them in a sophisticated manner to produce a higher-quality image. Therefore, burst photography allows inexpensive hardware to overcome mechanical and physical constraints and thus achieving higher imaging quality in the expense of computation time. While ideally, we would like each frame of the burst to capture precisely the same scene, this is not possible due to camera motion (e.g. hand shake), scene motion by dynamic moving objects and finally inefficiencies of Optical Image Stabilization (OIS) hardware that may cause a slight drift even for completely static scenes. Therefore, homography estimation and alignment usually is necessary when processing frames of the same scene.

The idea of using a sequence of photographs to enhance the image quality, is not new and it has been successfully exploited in the past for the tasks of image debluring~\cite{Aittala_2018_ECCV,jian.2009}, denoising~\cite{maggioni2011video} and super-resolution~\cite{farsiu.2004}. Inspired from these works, we design a restoration algorithm that involves a neural network, to handle various tasks of burst photography. First, we rely on a physical model for the observations of the burst, which in turn enables us to derive an optimization scheme for restoration purposes. The optimization scheme is combined with supervised learning of a neural network with a transparent architecture, leading to an Iterative Neural Network (INN). The developed framework exhibits by design many desired properties, which competing deep learning methods for burst photography do not necessarily exhibit, namely a) inherent invariance to the ordering of the frames, b) support of bursts of arbitrary size and c) scalability to burst size. 
\vspace{-.1cm}
\section{Related work}
\label{sec:Related_Work}
\vspace{-.1cm}
\subsection{Image Denoising}
Single image denoising is a longstanding problem, and it has progressed dramatically in recent decades, approaching its believed performance limit~\cite{levin.2011}.
The list of methods includes but not limited to Field-of-Experts~\cite{roth.2005.fields}, Non-Local Means~\cite{buades2005non} and BM3D~\cite{dabov2007image}, with the latter being the de facto method used till today. With the advent of deep learning, several learning-based methods have emerged during the last few years that take advantage of neural networks in order to push the reconstruction quality even further. Systems like DnCNN~\cite{DCNN}, NLNet~\cite{Lefkimmiatis_2017_CVPR} and MemNet~\cite{Tai-MemNet-2017} have succeeded to set a new state-of-the art performance for the image denoising task. Unfortunately, recent works empirically indicate that we are now close to the believed performance limit for single image denoising task, since quantitative performance improvements are no longer substantial and do not fully justify the simultaneous disproportionate increase of computational complexity.

Fortunately, burst denoising still allows the development of methods that can achieve better reconstruction than single image denoising.
In fact, several multi-frame variants of single-image denoising methods have been successfully developed. For example, VBM3D~\cite{kostadin2007video} and VBM4D~\cite{maggioni2011video} are two known extensions of the BM3D framework that work on videos and bursts of images, respectively. Furthermore, techniques as in~\cite{liu2014burstdenoising} were developed specifically for low resource photography applications and denoise an image using a burst sequence, in a fraction of the time required by VBM4D and other variants. Finally, modern deep learning approaches for burst denoising have recently emerged, such as those in~\cite{godard2017deep, mildenhall2018burst}, and provide insights for the success of end-to-end methods by achieving superior reconstruction quality. 

\subsection{Image Demosaicking}
While the literature on multi-image demosaicking methods falls short, demosaicking as a standalone problem has been studied for decades and for a complete survey we refer to~\cite{li.2008}. A very common approach is bilinear interpolation, as well as, other variants of this method which are adaptive to image edges~\cite{Hirakawa.2005, menon.2009}. During the last years, the image demosaicking task witnessed an incredible quantitative and qualitative performance increase via the use of neural network approaches like those in~\cite{Gharbi:2016:DJD:2980179.2982399, Henz.2018} and most recently in~\cite{kokkinos2018iterative}. This performance increase holds true even under the presence of noise perturbing the camera sensor readings.

Related to multi-frame photography, two well known systems that support burst demosaicking are FlexISP~\cite{heide2014flexisp} and ProxImaL~\cite{heide2016proximal}, which offer end-to-end formulations and joint solution via efficient optimization for many image processing related problems. Finally, a very successful commercial application on burst photography reconstruction is HDR+, introduced in~\cite{hasinoff2016burst}, where a burst of frames is utilized to alleviate shortcomings of smartphone cameras such as low dynamic range and noise perturbations. 

\section{Problem formulation}
\label{sec:Problem_Formulation}
To solve a variety of burst photography problems, we rely on the following observation model for each frame $\m y_i$ of a burst sequence of total size $B$,
\begin{equation}
\label{eq:model}
\m y_i = \m H S_i(\m x)+ \m n_i, \, i = 1\ldots, B.
\end{equation}
In Eq.~\eqref{eq:model}, $\m y_i \in\R^N$ corresponds to the degraded version of the affinely transformed underlying image $\m x \in \R^N$, which we aim to restore. While $\m x$ and $\m y_i$ are two dimensional images, for the sake of mathematical derivations, we assume that they have been raster scanned using a lexicographical order, and they correspond to vectors of $N$ dimensions. The operator $S_i: \R^N \xrightarrow{}\R^N$ is responsible for the affine transformation of the coordinate system of $\m x$. Specifically, it provides a mapping by interpolating values for each frame $i$ from the grid of the original image $\m x$. In our proposed method, we restrict the affine transformations to be rotational and translation so as to be on par with realistic burst photography applications. While in the above model it is assumed that the affine transformation is known, in practice we can only estimate it from the observations $\m y_i$ by setting one observation as a reference and aligning all other observations to the reference. This reference frame is considered to be completely aligned to the underlying image $\m x$,  and their relationship is described as $\m y_{ref} = \m H \m x + \m n_{ref}$.
Additionally, the underlying image  $\m x$ is further distorted by a linear operator $\m H\in\R^{N \times N}$, which describes a specific restoration problem that we aim to solve. This formulation is one of the most frequently used in the literature to model a variety of restoration problems such as image inpainting, deconvolution, demosaicking, denoising, and super-resolution. Each observation $\m y_i$ is also distorted by noise $\m n_i \sim \mathcal{N}(0,\,\sigma^2)$, which is assumed to follow an i.i.d Gaussian distribution.  

Recovering $\m x$ from the measurements $\m y_i$ belongs to the broad class of inverse problems. For most practical problems, the operator $\m H$ is typically singular, i.e., not invertible. This fact, coupled with the presence of noise perturbing the measurements and the affine transformation leads to an ill-posed problem where a unique solution does not exist. In general, such problems can be addressed following a variational approach. Under this framework, a solution is obtained by  minimizing an objective function of the form:
\vspace{-.2cm}
\begin{equation}
\label{eq:variational}
\m x^{\star} =\underbrace{\argmin_{\m x} \frac{1}{2\sigma^2B}\sum_{i=1}^{B}\norm{\m y_i-\m H S_i(\m x)}{2}^2}_{f\pr{\m x}} + r(\m x),
\vspace{-.1cm}
\end{equation}
where the first term corresponds to the data fidelity term that quantifies the proximity of the solution to the observations and the second term corresponds to the regularizer of the solution, which encodes any available prior knowledge we might have about the underlying image. As it can be seen from Eq.~\eqref{eq:variational}, the solution $\m x^{\star}$ must obey the observation model for each frame $\m y_i$ of the burst. While the above variational formulation is general enough to accommodate for a variety of different inverse problems, in Section~\ref{sec:Experiments} we focus on two particular problems: 1) joint demosaicking and denoising and 2) burst Gaussian denoising. In the first case, $\m H$ becomes a binary diagonal matrix that corresponds to the Color Filter Array (CFA) of the camera, while in the second case  $\m H$ reduces to the identity operator. %

As we mentioned earlier, the role of the regularizer is to promote solutions that follow specific image properties and as a result its choice significantly affects the end-result of the restoration. Some typical choices for regularizing inverse problems is the Total Variation~\cite{rudin1992nonlinear} and the Tikhonov~\cite{Tikhonov:1963} functionals. While such regularizers have been frequently used in the past in image processing and computer vision applications, their efficacy is limited. For this reason, in this work, we follow a different path, and we attempt to learn the regularizer implicitly from available training data. Therefore, throughout this work, we do not make any assumptions regarding the explicit form of the regularizer. Rather, as we will explain later in detail, our goal is to learn the effect of the regularizer to the solution through the proximal map~\cite{parikh2014proximal}.

\section{Proximal gradient descent}
\label{sec:PGD}
Efficiently solving Eq.~\eqref{eq:variational} has been a longstanding problem, and as a result a variety of sophisticated optimization methods have been proposed over the years. In our work, we employ a relatively simple method that extends the classical gradient descent, namely the Proximal Gradient Descent (PGD)~\cite{parikh2014proximal}. In particular, PGD is a generalization of gradient descent that can deal with the optimization of functions that are not fully differentiable but they can be split into a differentiable and a non-differentiable part, i.e. $F\pr{\m x} = s\pr{\m x} + g\pr{\m x}$. Then, according to PGD, the solution can be obtained in an iterative fashion as follows:
\begin{equation}
    \label{eq:proximal_gradient_descent}
    \m x^t = \prox_{\gamma g}(\m x ^{t-1} - \gamma \nabla_{\m x} s(\m x^{t-1})),
\end{equation}
where $\gamma$ is the step size and $\prox_{\gamma g}$ is the proximal operator, related to the non-smooth part of the overall function, $g(\m x)$, and the step size $\gamma$. Typically, $\gamma$ is adaptive and is computed using a line-search algorithm. However, when $s\pr{\cdot}$ is Lipschitz continuous it can be fixed and set as $\gamma = \frac{1}{L}$, where $L$ is the Lipschitz constant of $\grad_{\m x}s$. In each iteration $t$, first a gradient descent step is performed for the smooth part $s\pr{\m x}$ of the objective function, while in the sequel the non-smooth term is handled via the proximal operator, whose action on a vector $\m v$ is defined as:
\begin{equation}
    \label{eq:proximal_operator}
    \prox_{\gamma g}(\m v) = \argmin_{\m z} \tfrac{1}{2} \norm{\m v - \m z}{2}^2 + \gamma g(\m z).
    \vspace{-.1cm}
\end{equation}

From a signal processing perspective, the proximal map corresponds to the regularized solution of a Gaussian denoising problem, where $\m v$ is the noisy observation, $g\pr{\cdot}$ is the employed regularizer and $\gamma$ the regularization parameter. Based on the above and by inspecting Eq.~\eqref{eq:variational}, we observe that in our case the data fidelity corresponds to the smooth part while we further consider the regularizer as the non-smooth part. We note, the most effective regularizers in variational methods have been shown to be indeed non-differentiable and, thus, our assumption is a reasonable one.

Referring to Eq.~\eqref{eq:variational}, the gradient of the data fidelity term can be easily computed as  \vspace{-.1cm}
\begin{equation}
\label{eq:gradient_of_data_fidelity}
    \nabla_{\m x} f(\m x) = \frac{1}{\sigma^2B}\sum_{i=1}^{B} \nabla_{\m x} S_i(\m x) \m H^\transp (-\m y_i + \m H S_i(\m x)).
    \vspace{-.2cm}
\end{equation}
A useful observation is that the gradient of $f\pr{\m x}$ can be linearized and therefore the time-consuming calculation of the Jacobian of the affine transform $S_i$ can be entirely avoided. The base of this observation is that the mapping $S_i(\m x)$ corresponds to an interpolation, such as bilinear, on an image $\m x$ with respect to a certain warping matrix. By calculating beforehand the new pixel locations, using the estimated warping matrix, that we would like to interpolate from the image $\m x$, the interpolation itself can be re-written as a linear operation $\m S_i\m x$. In this case, $\m S_i$ is a sparse matrix with only a few of its columns being non-zero and which hold the coefficients for the weighted averaging of pixel intensities. Therefore, under this approach it holds that $\m S_i \m x = S_i(\m x)$. For example, in the case of bilinear interpolation only four elements of each row of the matrix $\m S_i$ will be non-zero, while in the case of nearest neighbor interpolation only one element is non-zero and is equal to one. 

Consequently, the gradient of the data fidelity term can be rewritten as
\vspace{-.2cm}
\begin{equation}
\label{eq:linearized_gradient_of_data_fidelity}
    \nabla_{\m x} f(\m x) = \frac{1}{\sigma^2B}\sum_{i=1}^{B} \m S_i^\transp \m H^\transp (-\m y_i + \m H \m S_i\m x),
    \vspace{-.2cm}
\end{equation}
where $\m S_i^\transp$ is the adjoint operator of $\m S_i$. This adjoint operation amounts to interpolating an image $\m x$ with the inverse of the warping matrix. In our case, this matrix is always existent, since we have restricted our affine transformations to support only rotation and translation. Finally, by using the gradient of the data fidelity term of Eq.~\eqref{eq:linearized_gradient_of_data_fidelity}, in the proximal gradient step described in ~Eq.~\eqref{eq:proximal_gradient_descent}, and by computing its Lipschitz constant as $L=\tfrac{1}{\sigma^2}$ (the proof is provided in the supplementary material), we end up with the following iterative optimization step for burst photography applications
\vspace{-1cm}

\begin{equation}
     \label{eq:proximal_gradient_descent_full}
     \m x^t = \prox_{\sigma^2 r}(\m x ^{t-1}\!+\!\frac{1}{B}\sum_{i=1}^{B} \m S_i^\transp \m H^\transp (\m y_i - \m H \m S_i\m x^{t-1})).
    \vspace{-.2cm}
 \end{equation}

In order to retrieve the solution of the minimization problem in Eq.~\eqref{eq:variational} based on the above iterative scheme, the appropriate form of $r(\m x)$ must be first specified. However, this is far from a detrimental task. Apart from this, the convergence to a solution usually requires a large number of iterations, which implies a significant computational cost.

To deal with these challenges, in this work we pursue a different approach than conventional regularization methods. In particular, instead of selecting a specific regularizer and deriving the solution via Eq.~\eqref{eq:proximal_gradient_descent_full}, we design a network to learn the mapping between the proximal input and the denoised output. This strategy allows us to unroll K iterations of the PGD method and use a suitable network to approximate the output of the proximal operator. It is important to note that this approach does not carry any risk of leading to a reconstruction of inferior quality. The reason is that in large scale optimization techniques, even when the regularizer is fully specified, typically the proximal map cannot be computed in closed-form. In such cases~\cite{beck2009fast,lefkimmiatis.2013}, it is roughly approximated via iterative schemes without this jeopardizing the overall reconstruction quality.%
 Another important point we would like to highlight is that our approach, as opposed to other related methods that use a network to replace the proximal operator such as IRCNN~\cite{zhang.2017}, Plug and Play~\cite{Venkatakrishnan.2013} and RED~\cite{romano2017little}, is completely parameter-free and, thus, no manual tuning is required so that a good reconstruction is produced.
\vspace{-0.1cm}
\section{Proposed Iterative Neural Network (INN)}
\label{sec:Iterative_Neural_Network}
\subsection{Proximal Network}
As described in Section~\ref{sec:PGD}, the proximal map can be interpreted as the regularized solution of a Gaussian denoising problem. Based on this observation, we can exploit the capabilities of neural networks and replace the iterative computation of the proximal map with a CNN that takes as input a noisy image and the standard deviation of the noise and returns as output the denoised version of the input. 

While there are many image denoising neural networks such as the DnCNN \cite{DCNN} or MemNet\cite{Tai-MemNet-2017} that we could use to approximate the proximal map, in this work we employ the ResDNet network described in~\cite{Kokkinos_2018_ECCV}, which was originally inspired by  UDNet~\cite{lefkimmiatis.2017}. Similarly to DnCNN, ResDNet is a fully convolutional denoising network and can handle a wide range of noise levels by using a single set of parameters. It also has a residual architecture since instead of estimating directly the denoised image, it first estimates a noise realization which is then subtracted from the noisy input. The advantage of ResDNet over DnCNN is that it takes as an additional input the standard deviation of the noise, which is then used by the network to normalize the noise estimate in order to ensure that it has the desired variance. This feature is instrumental for the successful implementation of our overall scheme, as it allows us to have more control over the output of the network.

In detail, the architecture of ResDNet consists of $D$ residual blocks with 2 convolutional layers each of 64 filters and kernels with dimensionality $3\times3$. The residual blocks precedes a convolutional layer applied on the input which increases the number of channels from 3 to 64 using kernels of size $5\times5$. The feature maps are eventually decreased to 3 from 64 via a convolutional layer with a kernel of support $5\times5$. In every step, the employed non-linearity, which is applied after every convolutional layer, except the last one, is the parametrized rectified linear unit (PReLU)~\cite{He_2015_ICCV}. The end result of ResDNet is a noise realization estimate that is subtracted from the distorted image. Before the subtraction takes place, the noise realization is normalized so that its variance matches the input variance. This is accomplished with a trainable $\ell_2$-projection layer, 
\begin{equation}
\Pi_{\mathcal{C}}\pr{\m y} =\theta \m y /\max(\norm{\m y}{2},\theta),
\end{equation}
where $\theta=\sigma\sqrt{N-1}$. Overall, this denoising network is relatively small since it contains approximately 380K parameters and it can be easily deployed in each iteration of our INN without requiring excessive memory or computation time. 

\newlength{\textfloatsepsave} 
\setlength{\textfloatsepsave}{\textfloatsep} 
\setlength{\textfloatsep}{0pt}

\begin{algorithm}[h]
 \SetAlgoCaptionSeparator{\unskip:}
 \SetKwInOut{Input}{Input}
 \Input{$\m H$: Degradation Operator, $\m y_{\{1...B\}}$: input burst, $K$: iterations, $\bm w\in\R^K$: extrapolation weights, $\sigma$: estimated noise, $\bm s\in\R^K$: projection parameters}
 $\m x^{0}= \m 0$\;
 Initialize $\m x^{1}$ using $\m y_{ref}$\;
 Estimate mappings $\m S_{1...B}$\;
 
  \For{$t\gets1$ \KwTo $K$}{
    $ \m u = \m x ^{t} + \bm w_t (\m x^{t} - \m x^{t-1}) $\;
    $ \m {z} = 0 $\;
    \For{$i\gets1$ \KwTo $B$}{
      $ \m {z} = \m {z} + \m S_i^\transp \m H^\transp (-\m y_i + \m H \m S_i\m u) $\;
   }
    $\m x^{t+1} = \text{ProxNet}(\m x^t- \m {z}/B, \sigma, \bm s_t )$\; 
 }
 \caption{Proposed Iterative Neural Network for bust photography applications}
  \label{alg:the_alg}
\end{algorithm}
In order to emphasize that the employed denoising network in our INN serves as a proximal map estimate and not as a single image Gaussian denoiser, hereafter, we will refer to it as ProxNet. Another reason for our naming convention is that our overall approach is not tied to a specific proximal network and in principle ResDNet can be replaced by another network architecture that exhibits similar properties.   
\setlength{\textfloatsep}{\textfloatsepsave}

\subsection{Iterative neural network}
The proposed INN combines the PGD algorithm as discussed in Section~\ref{sec:PGD} and the proximal network as an estimator of the solution of Eq.~\eqref{eq:proximal_operator}. A straightforward way to implement the INN is to use in every iteration a proximal network that is governed by a different set of parameters. However, the training of INN, in this case, becomes quickly intractable, since the number of parameters increases linearly to the number of employed iterations. %
To deal with this shortcoming, we instead use the same proximal network in every iteration, and thus we keep the number of network parameters small, which in turn decreases the necessary training time and the memory footprint of the network. 

In order to speed up the convergence of the optimization scheme, we exploit two commonly used convergence acceleration strategies. The first one is the homotopy continuation strategy~\cite{Lin2015} where the standard deviation of the noise is deliberately over-estimated in the first iterations and gradually is decreased until the accurate estimation of $\sigma$ is reached. The homotopy continuation scheme accelerates the convergence of PGD algorithms as shown in~\cite{Lin2015} and it can be easily integrated into our formulation via a modification of the projection layer by replacing $\theta$ with $\hat{\theta}=e^s \theta$. In detail, we initialize the trainable parameter of the projection layer $\bm s \in \mathbb{R}^K$ with values spaced evenly on a log scale from $s_{max}$ to $s_{min}$ and later on the vector $\bm s$ is further finetuned on the training dataset via back-propagation. 

The second acceleration strategy that we explore involves the use of an extrapolation step similar to the one introduced in~\cite{fista.2009}. Specifically, the outputs of two consecutive iterations are combined in a weighted manner in order to obtain the solution of the current iteration. In~\cite{fista.2009} the extrapolation weights $\m w \in \R^K$ are known apriori but in our work, we learn them during the training of INN. %
We initialize the extrapolation weights as $w_i = \frac{t-1}{t+2}\text{,} \forall 1 \le t \le K$, which matches the configuration described in~\cite{parikh2014proximal}.

Algorithm~\ref{alg:the_alg} describes our overall strategy which combines all the different components that we described, i.e., the PGD, the proximal network, the continuation, and extrapolation strategies. 
As it can be seen from Algorithm~\ref{alg:the_alg}, our reconstruction approach has only a weak dependency on the burst size, since this is only involved in the computation of the gradients for each burst observation, which can be done very efficiently. This feature makes our method very efficient since the proximal network is independent to the bust size $B$, unlike other recent deep learning based methods~ \cite{godard2017deep,Aittala_2018_ECCV}, which process each frame of the burst individually at first and then jointly and therefore the computational time increases linearly to $B$. Simultaneously, our proposed approach supports by design bursts of arbitrary size with only a minor computational overhead. We note that this is not the case for the network in~\cite{mildenhall2018burst} which is constrained to use bursts of 8 frames. In a different case, the entire network needs to be trained from scratch. Finally, our proposed INN is by definition permutation invariant similarly to~\cite{Aittala_2018_ECCV}. In particular, the ordering of the burst frames does not affect at all the reconstruction result as long as the reference frame remains the same. 

\section{Network Training}
\subsection{Synthetic training dataset}
Since there are no publicly available burst photography datasets suitable for training our network, we create training pairs of ground-truth and input bursts using the Microsoft Demosaicking Dataset (MSR)~\cite{khasabi2014} for burst image demosaicking and the Waterloo Dataset~\cite{ma2017waterloo} for burst Gaussian denoising. In both cases, we modify the ground-truth image by affinely transforming it 8 times to create a burst with synthetic misalignment and then the images are center cropped to retain a patch of $128 \times 128$ pixel. We assume that the reference frame is the last one and therefore it does not undergo any transformation. The random affine transformation should be close to realistic scenarios, and thus we restrict the transformation to contain a translation in each direction up to 10 pixels and rotation of up to 2 degrees. %

For burst image demosaicking, we selected the MSR dataset which is a small but well-known dataset for evaluating image demosaicking algorithms, as explained in~\cite{khasabi2014}. The advantage of the MSR dataset is that all data are in the linear color space where pixel measurements are proportional to the number of counted photons, and no post-processing steps have been performed (e.g., sharpening, tone mapping) that will alter the image statistics. The dataset consists of 200 images for training, 100 for validation and 200 images for testing purposes. For each ground-truth image we generate the respective burst sequence, and then we apply the Bayer pattern on each frame. We also explore the case of noise perturbing the camera measurements, and therefore we add noise sampled from a heteroskedastic Gaussian distribution with signal dependent standard deviation  $\hat {\m \omega} \sim \mathcal{N}(\m \omega, \alpha \m \omega + \beta^2)$, following the model presented in~\cite{Healey.1994}. The parameter $\alpha$ is related to the shot noise component, which occurs from the stochastic nature of the photon counting process and it is dependent on the true intensities $\m y$, while the parameter $\beta$ is linked to the signal independent read noise component. Both noise parameters are sampled uniformly from a specific range as discussed in~\cite{mildenhall2018burst}, which covers the noise levels of many widely used cameras. The dataset is also augmented with random flipping and color jittering in order to ensure a plethora of lighting conditions.

For burst image denoising, we use the Waterloo dataset which consists of 4,744 images. Using the described procedure, we retrieved the synthetically mis-aligned bursts of 8 frames and 500 of these bursts were kept separately to be used for testing purposes. All frames were distorted with additive Gaussian noise with standard deviation sampled from $[5,25]$ with a step size equal to $2.5$. 

For all experiments, we estimate the warping matrix that aligns every observation to the reference frame using the Enhanced Correlation Coefficient (ECC)~\cite{Evangelidis.2008}. Since the images are severely distorted by noise, we estimate the alignment on the Gaussian pyramid of the image and use the warping matrix of coarse scales to initialize the ECC estimation of finer levels in order to achieve robustness to the noise perturbations. Bursts that failed to be aligned using this method were dropped from the training set. 

\subsection{Implementation Details}
For all experiments we choose the interpolation operation, involved in the affine transformation of the observation model Eq.~\eqref{eq:model}, to be bilinear due to its low computation complexity and the adequate result that it provides. Using a pre-trained proximal network our overall network is further trained end-to-end to minimize the $\ell_1$ loss.

Due to the iterative nature of our framework, the network parameters are updated using the Back-Propagation Through Time (BPTT) algorithm, and more specifically we adopt the Truncated BPTT framework presented in~\cite{robinson.2008, kokkinos2018iterative}. While we unfold $K$ instances of the network, we propagate the gradients through smaller chunks of size $k$ instead of $K$, due to the inherent memory restrictions we face during training. Every $k$ iterations we update the parameters based on the loss function and then proceed with unrolling the next $k$ iterations till the number of total iterations $K$ is reached. This modification of the standard BPTT allows the usage of larger batch sizes and a higher number of iterations which leads to better performance, as shown in~\cite{kokkinos2018iterative}. 
Furthermore, we set for all experiments $K=10$, $k=5$ and the optimization is carried via the AMSGRAD optimizer where the training starts from an initial learning rate which we decrease by a factor of 10 every 100 epochs. %
The specific hyper-parameters used for training of each model are provided in the supplementary material.

\section{Experiments}
\label{sec:Experiments}
\subsection{Image Demosaicking and Denoising}
\label{sec:experiment_demosaicking}

\begin{table}[t]
\centering
\begin{tabular}{|l|c|c|c|c|}
\hline
 & \multicolumn{2}{c|}{noisy} & \multicolumn{2}{c|}{noise-free} \\ \hline
 & linRGB & sRGB & linRGB & sRGB \\ \hline
\multicolumn{5}{|l|}{Bilinear} \\ \hline
\multicolumn{1}{|c|}{- single} & 27.62 & 23.02 & 29.07 & 22.86 \\ \hline
\multicolumn{1}{|c|}{- burst} & 30.03 & 26.45 & 31.46 & 27.23 \\ \hline
\multicolumn{5}{|l|}{Gharbi \cite{Gharbi:2016:DJD:2980179.2982399}} \\ \hline
\multicolumn{1}{|c|}{- single} & 36.52 & 31.37 & 41.08 & 34.46 \\ \hline
\multicolumn{1}{|c|}{- burst} & 37.14 & 31.87 & 39.74 & 34.39 \\ \hline
\multicolumn{5}{|l|}{Kokkinos \cite{kokkinos2018iterative}} \\ \hline
\multicolumn{1}{|c|}{- single} & 38.48 & 33.41 & 41.03 & 34.37 \\ \hline
\multicolumn{1}{|c|}{- burst} & 38.06 & 33.06 & 38.93 & 33.02 \\ \hline
\multicolumn{5}{|l|}{BM3D-CFA\cite{Danielyan_2009_BM3DCFA}} \\ \hline
\multicolumn{1}{|c|}{- single} & 35.63 & 30.49 & - & - \\ \hline
\multicolumn{1}{|c|}{- burst} & 35.36 & 30.30 & - & - \\ \hline
\hline
Ours & \textbf{39.64} & \textbf{34.56} & \textbf{42.40} & \textbf{36.24} \\ \hline
Ours (oracle) & 41.55 & 35.59 & 42.40 & 36.24 \\ \hline
\end{tabular}
\caption{PSNR performance of different methods in both linear and sRGB spaces. Every method was tested on both single image and burst scenario. In the case of BM3D-CFA, demosaicking of the denoised images was performed using the noisefree model of~\cite{kokkinos2018iterative}.\vspace{-0.5cm}}
\label{tbl:msr}
\end{table}

\begin{figure*}[t]
  \centering
  \subfloat[Ground Truth]{\includegraphics[width=0.125\textwidth,valign=c]{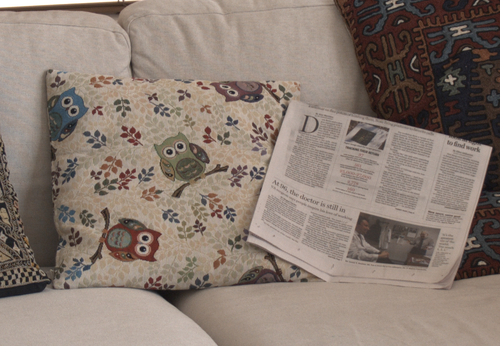}}
  \subfloat[Input]{\includegraphics[width=0.125\textwidth,valign=c]{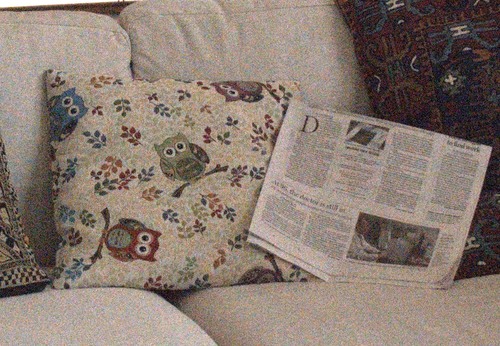}}
  \subfloat[Ours]{\includegraphics[width=0.125\textwidth,valign=c]{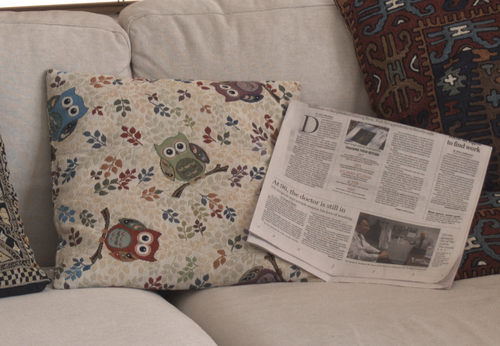}}
  \subfloat[HDR+~\cite{hasinoff2016burst}]{\includegraphics[width=0.125\textwidth,valign=c]{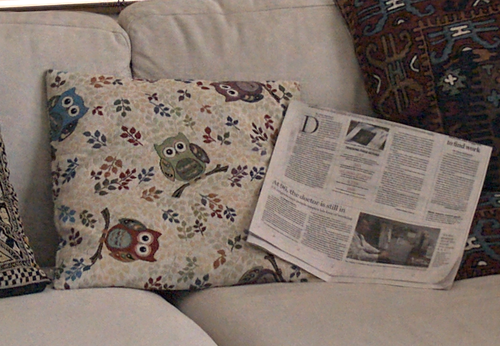}}
  \subfloat[FlexISP~\cite{heide2014flexisp}]{\includegraphics[width=0.125\textwidth,valign=c]{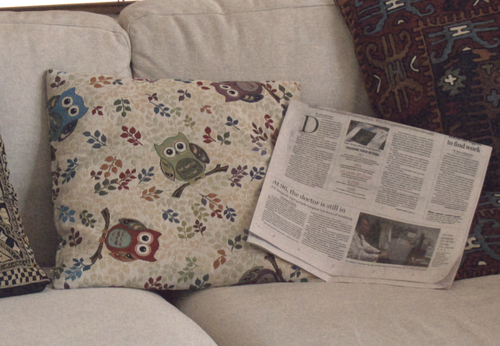}}
  \subfloat[Godard~\cite{godard2017deep}]{\includegraphics[width=0.125\textwidth,valign=c]{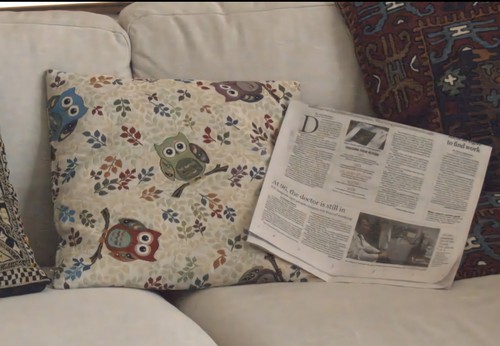}}
  \subfloat[Gharbi~\cite{Gharbi:2016:DJD:2980179.2982399}]{\includegraphics[width=0.125\textwidth,valign=c]{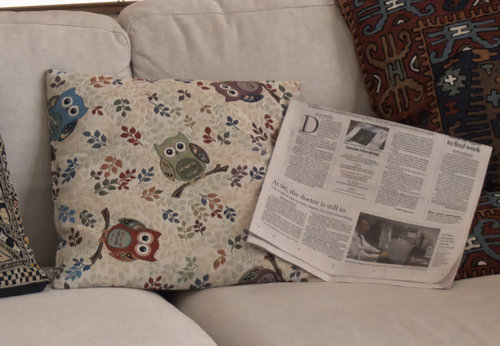}}
  \subfloat[Kokkinos~\cite{Kokkinos_2018_ECCV}]{\includegraphics[width=0.125\textwidth,valign=c]{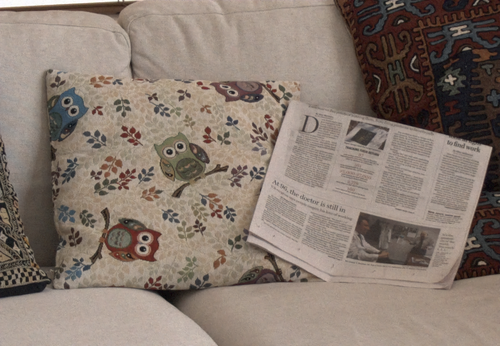}}
  \\
  \vspace{-0.35cm}
  \subfloat{\subfloat {\includegraphics[width=0.0625\textwidth,valign=c]{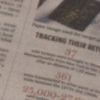}}
                          \subfloat {\includegraphics[width=0.0625\textwidth,valign=c]{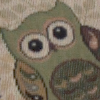}}}
  \subfloat{\subfloat {\includegraphics[width=0.0625\textwidth,valign=c]{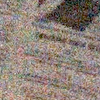}}
                          \subfloat {\includegraphics[width=0.0625\textwidth,valign=c]{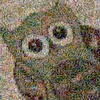}}}
  \subfloat{\subfloat{\includegraphics[width=0.0625\textwidth,valign=c]{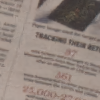}}
            \subfloat {\includegraphics[width=0.0625\textwidth,valign=c]{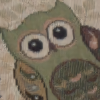}}}
  \subfloat{\subfloat {\includegraphics[width=0.0625\textwidth,valign=c]{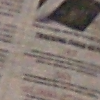}}
                          \subfloat {\includegraphics[width=0.0625\textwidth,valign=c]{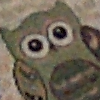}}}
  \subfloat{\subfloat{\includegraphics[width=0.0625\textwidth,valign=c]{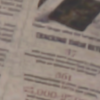}}
                          \subfloat {\includegraphics[width=0.0625\textwidth,valign=c]{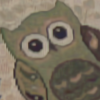}}}
  \subfloat{\subfloat {\includegraphics[width=0.0625\textwidth,valign=c]{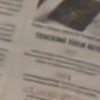}}
                          \subfloat {\includegraphics[width=0.0625\textwidth,valign=c]{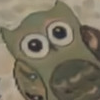}}}
  \subfloat{\subfloat{\includegraphics[width=0.0625\textwidth,valign=c]{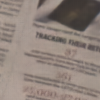}}
                          \subfloat {\includegraphics[width=0.0625\textwidth,valign=c]{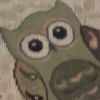}}}
  \subfloat{\subfloat{\includegraphics[width=0.0625\textwidth,valign=c]{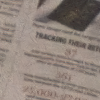}}                   \subfloat {\includegraphics[width=0.0625\textwidth,valign=c]{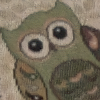}}}
  \\
  \vspace{-0.7cm}
  \subfloat{\includegraphics[width=0.125\textwidth,valign=c]{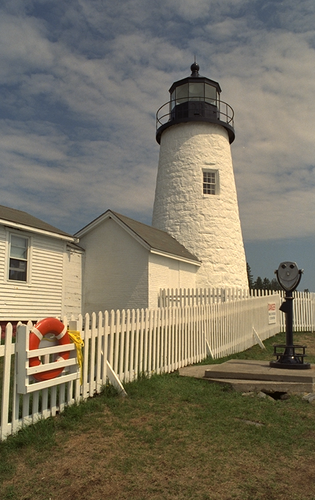}}
  \subfloat{\includegraphics[width=0.125\textwidth,valign=c]{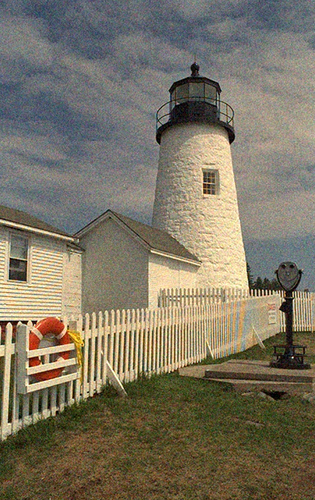}}
  \subfloat{\includegraphics[width=0.125\textwidth,valign=c]{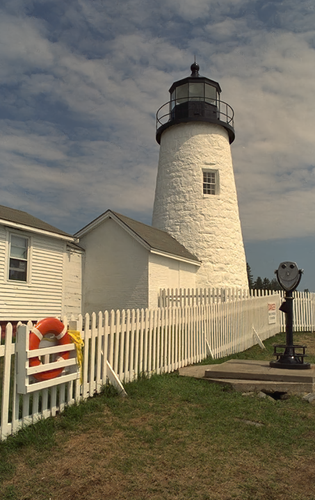}}
  \subfloat{\includegraphics[width=0.125\textwidth,valign=c]{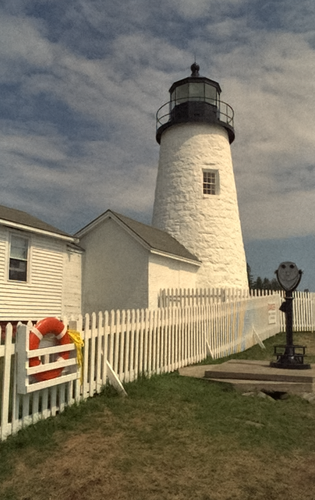}}
  \subfloat{\includegraphics[width=0.125\textwidth,valign=c]{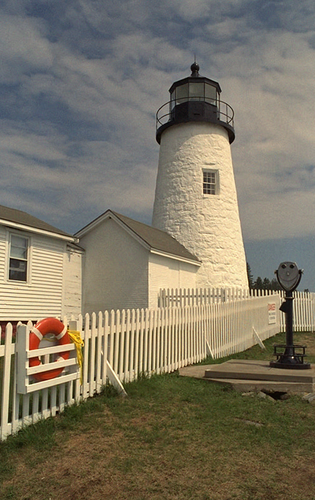}}
  \subfloat{\includegraphics[width=0.125\textwidth,valign=c]{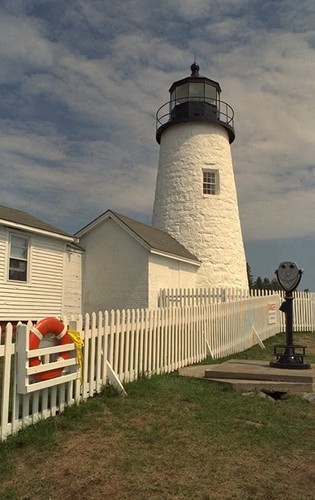}}
  \subfloat{\includegraphics[width=0.125\textwidth,valign=c]{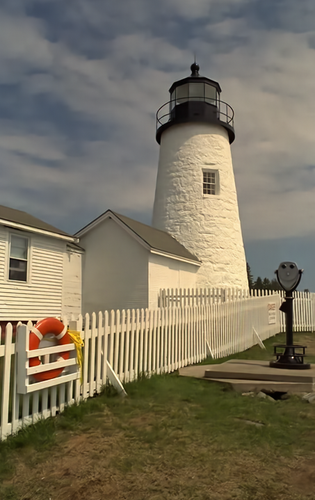}}
  \subfloat{\includegraphics[width=0.125\textwidth,valign=c]{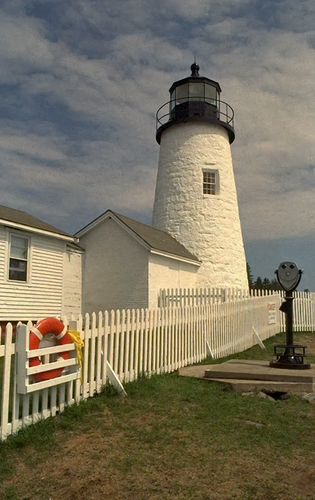}}
  \\
  \vspace{-0.35cm}
  \subfloat{\subfloat {\includegraphics[width=0.0625\textwidth,valign=c]{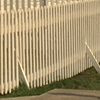}}
                          \subfloat {\includegraphics[width=0.0625\textwidth,valign=c]{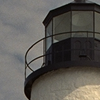}}}
  \subfloat{\subfloat {\includegraphics[width=0.0625\textwidth,valign=c]{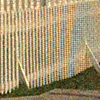}}
                          \subfloat {\includegraphics[width=0.0625\textwidth,valign=c]{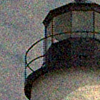}}}
  \subfloat{\subfloat {\includegraphics[width=0.0625\textwidth,valign=c]{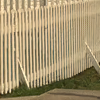}}
                          \subfloat {\includegraphics[width=0.0625\textwidth,valign=c]{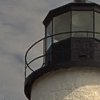}}}
  \subfloat{\subfloat {\includegraphics[width=0.0625\textwidth,valign=c]{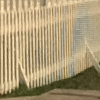}}
                          \subfloat {\includegraphics[width=0.0625\textwidth,valign=c]{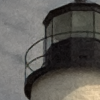}}}
  \subfloat{\subfloat {\includegraphics[width=0.0625\textwidth,valign=c]{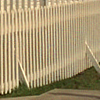}}
                          \subfloat {\includegraphics[width=0.0625\textwidth,valign=c]{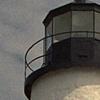}}}
  \subfloat{\subfloat {\includegraphics[width=0.0625\textwidth,valign=c]{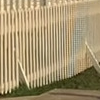}}
                          \subfloat {\includegraphics[width=0.0625\textwidth,valign=c]{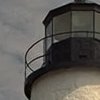}}}
  \subfloat{\subfloat {\includegraphics[width=0.0625\textwidth,valign=c]{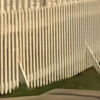}}
                          \subfloat {\includegraphics[width=0.0625\textwidth,valign=c]{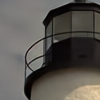}}}
  \subfloat{\subfloat {\includegraphics[width=0.0625\textwidth,valign=c]{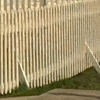}}
                          \subfloat {\includegraphics[width=0.0625\textwidth,valign=c]{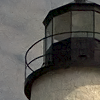}}}
  \vspace{-.35cm}
  \caption{Burst demosaicking results on a real and a synthetic burst from the FlexISP dataset~\cite{heide2014flexisp} (results are best seen magnified on a computer screen). Our model successfully restores the missing colors of the underlying images while suppressing noise. A PSNR comparison of the systems is provided in the supplementary material.} 
  \label{fig:burst_demosaicking}
\end{figure*}

We evaluate our method on the test set of the burst MSR dataset. In Table~\ref{tbl:msr}, we compare our INN with a bilinear interpolation baseline, two recent demosaicking neural networks~\cite{Gharbi:2016:DJD:2980179.2982399,kokkinos2018iterative}, as well as with a denoising approach using BM3D-CFA~\cite{Danielyan_2009_BM3DCFA} followed by demosaicking using the noise-free model of~\cite{kokkinos2018iterative}. BM3D-CFA was also used to denoise the raw data for the bilinear interpolation baseline in the noisy scenario. In all comparisons, we consider both a single image scenario and a burst variant where we apply the respective method on each frame of the burst and then the frames are aligned in order to be averaged. Our approach yields substantially better quantitative results than competing methods in both noisy and noise-free scenario with performance gains ranging from 0.9 to 1.5 dB. To visually assess the superiority of our approach, we further provide representative results in Fig.~\ref{fig:burst_demosaicking}.

In order to examine how the alignment of observations affects the results, we have also considered the case where our pre-trained network was fed with oracle warping matrices. As it could be expected, the restoration performance increases up to 1.9 dBs, which highlights the importance of robust image alignment and indicates that we can expect an increase in our network's performance by employing a better alignment method than the one we currently use.
\vspace{-0.1cm}
\subsection{Gaussian Image Denoising}

\label{sec:experiment_denoising}
We tested our method on the Gaussian denoising task where most burst photography methods focus on. For comparisons, we used the methods of BM3D, VBM4D and ResDNet for single and burst scenarios. In the case of the burst variant of ResDNet, the images were first denoised using ResDNet and then aligned using the method~\cite{Evangelidis.2008} before being averaged. For reasons of experimental completeness, we would like to compare our method against the two most recent deep learning approaches~\cite{godard2017deep, mildenhall2018burst}, however, neither one of the models or their respective testing sets are publicly available yet. From the results presented in Table~\ref{tbl:denoising} and Fig.~\ref{fig:burst_denoise}, it is clear that our method achieves a state-of-the art performance across every noise level. An interesting result is that our INN, which uses ResDNet as a sub-component, consistently outperforms the burst variant of ResDNet. This is attributed to the principled way we designed our INN so that it faithfully follows the forward model.

\begin{figure}[t]

\begin{center}

\includegraphics[width=1.\linewidth]{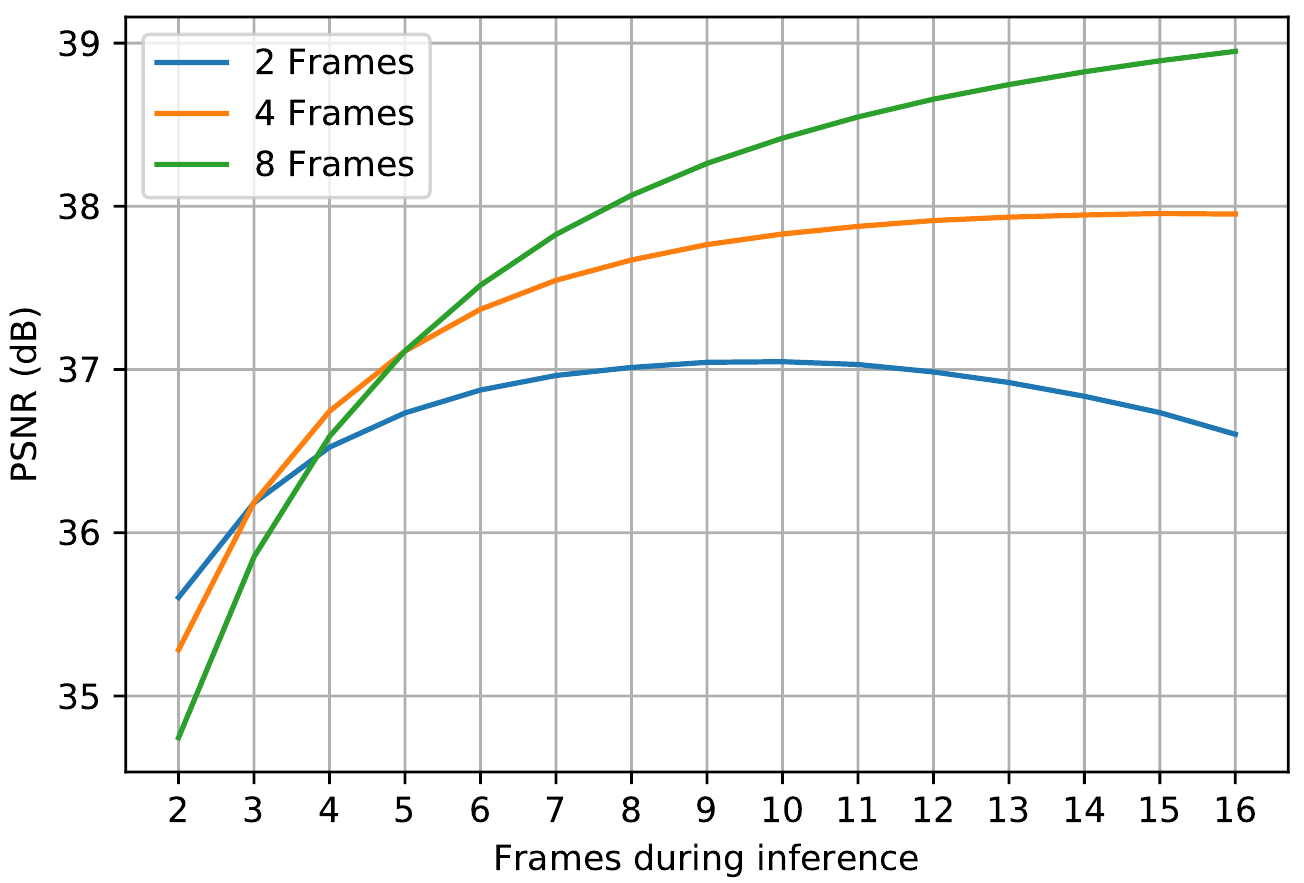}
\vspace{-1cm}\end{center}
   \caption{Generalization ability of our INN to different burst sizes. Three models were trained with 2, 4 and 8 frames and tested on burst sequences from 2 to 16 frames.\vspace{-0.5cm}}
\label{fig:frames_seq}
\end{figure}

We also performed an ablation study on the importance of burst size during training. Specifically,  we trained 3 models using bursts of size 2, 4 and 8 and tested them on sequences with burst sizes varying from 2 to 16, as presented in Fig.~\ref{fig:frames_seq}. The models that were trained with 4 and 8 frames are able to generalize well when they are provided with more frames during inference since their performance steadily increases. Nevertheless, there is a performance gap between the models which indicates that the burst size for which the network is originally trained for can affect the performance. The model trained to handle bursts of only two frames exhibits the same behaviour up to a certain number of frames but after that, its performance starts to decline. Our findings contradict the conclusion of the authors in~\cite{godard2017deep} that deep learning models need to be trained with many frames in order to generalize to longer sequences during inference. In fact, our network variants trained for 4 and 8 bursts show a consistent performance improvement with the increase of the burst sequence. 

\begin{table}[t]
\centering
\begin{tabular}{|c|c|c|c|c|c}
\hline
Methods & $\sigma$=5 & $\sigma$=10 & $\sigma$=15 & $\sigma$=20 & \multicolumn{1}{c|}{$\sigma$=25} \\ \hline
noisy ref. frame & 34.26 & 28.37 & 24.95 & 22.55 & \multicolumn{1}{c|}{20.71} \\ \hline
BM3D & 39.78 & 35.86 & 33.55 & 31.86 & \multicolumn{1}{c|}{30.50} \\ \hline
VBM4D & 39.64 & 35.67 & 33.35 & 31.67 & \multicolumn{1}{c|}{30.34} \\ \hline
\multicolumn{6}{|l|}{ResDNet:} \\ \hline
- single & \textbf{40.19} & 36.65 & 34.55 & 33.03 & \multicolumn{1}{c|}{31.82} \\ \hline
- burst & 39.69 & 37.65 & 36.06 & 34.89 & \multicolumn{1}{c|}{33.86} \\ \hline
\hline
Ours & 40.08 & \textbf{38.71} & \textbf{37.36} & \textbf{36.24} & \multicolumn{1}{c|}{\textbf{35.28}} \\ \hline
\end{tabular}
\caption{Color image denoising comparisons for five different noise levels. The restoration quality is measured in terms of average PSNR.}
\label{tbl:denoising}
\end{table}

\vspace{-0.1cm}
\section{Limitations}
\label{sec:Limitations}
Our method is capable of producing high-quality images from a burst sequence with great success. However, the main limitation of our network is the dependency it has to the ECC estimation of the warping matrix, which in practice can be rather inaccurate especially when there is a strong presence of noise. When the estimated affine transformation matrix is imprecise, our network inevitably will introduce ghosting artifacts to the final result Fig. \ref{fig:burst_denoise} (more examples can be found in the supplementary material). In this event, one possible solution, is to estimate the quality of the transformation matrix via a consistency metric like the one in~\cite{liu2014burstdenoising} and crop out inconsistent areas from a frame.  
\vspace{-0.1cm}
\section{Conclusions}
\label{sec:Conclusions}
In this work, we have proposed a novel iterative neural network architecture for burst photography applications. Our derived network has been designed to respect the physical model of burst photography while its overall structure is inspired by large-scale optimization techniques. By explicitly taking into account the special characteristics of the problems under study, our network outperforms previous state-of-the-art methods across various tasks, while being invariant to the ordering of the frames and capable to generalize well to arbitrary burst sizes.

\begin{figure*}[h]
  \centering
  \subfloat[Ground Truth]{\includegraphics[width=0.16\textwidth,height=0.13\textwidth]{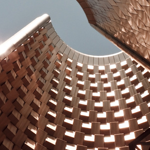}}
  \subfloat[Average]{\includegraphics[width=0.16\textwidth,height=0.13\textwidth]{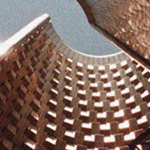}}
  \subfloat[Ours]{\includegraphics[width=0.16\textwidth,height=0.13\textwidth]{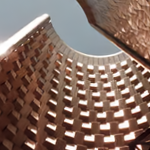}}
  \subfloat[ResDNet]{\includegraphics[width=0.16\textwidth,height=0.13\textwidth]{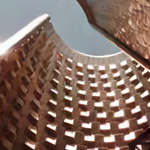}}
  \subfloat[ResDNet Average]{\includegraphics[width=0.16\textwidth,height=0.13\textwidth]{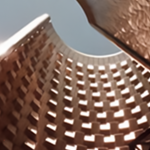}}
  \subfloat[VBM4D~\cite{maggioni2011video}]{\includegraphics[width=0.16\textwidth,height=0.13\textwidth]{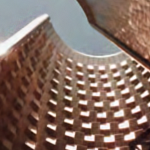}}\\
  \vspace{-0.35cm}
  \subfloat{\includegraphics[width=0.16\textwidth,height=0.13\textwidth]{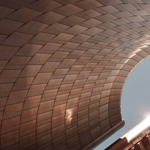}}
  \subfloat{\includegraphics[width=0.16\textwidth,height=0.13\textwidth]{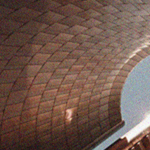}}
  \subfloat{\includegraphics[width=0.16\textwidth,height=0.13\textwidth]{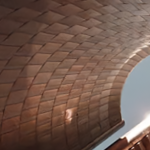}}
  \subfloat{\includegraphics[width=0.16\textwidth,height=0.13\textwidth]{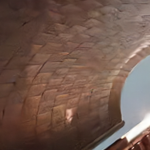}}
  \subfloat{\includegraphics[width=0.16\textwidth,height=0.13\textwidth]{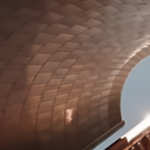}}
  \subfloat{\includegraphics[width=0.16\textwidth,height=0.13\textwidth]{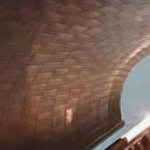}}\\
  \vspace{-0.35cm}
  \subfloat{\includegraphics[width=0.16\textwidth,height=0.13\textwidth]{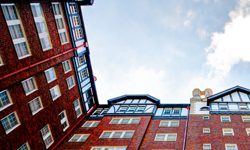}}
  \subfloat{\includegraphics[width=0.16\textwidth,height=0.13\textwidth]{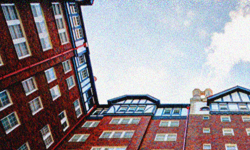}}
  \subfloat{\includegraphics[width=0.16\textwidth,height=0.13\textwidth]{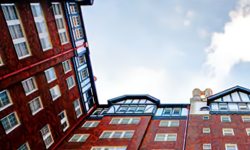}}
  \subfloat{\includegraphics[width=0.16\textwidth,height=0.13\textwidth]{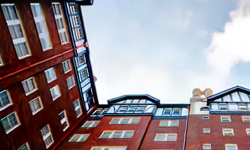}}
  \subfloat{\includegraphics[width=0.16\textwidth,height=0.13\textwidth]{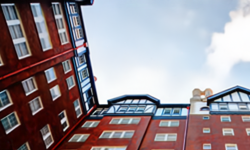}}
  \subfloat{\includegraphics[width=0.16\textwidth,height=0.13\textwidth]{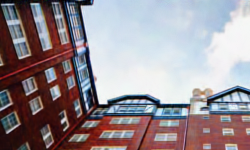}}\\
  \vspace{-0.35cm}
  \subfloat{\includegraphics[width=0.16\textwidth,height=0.13\textwidth]{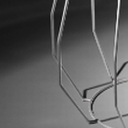}}
  \subfloat{\includegraphics[width=0.16\textwidth,height=0.13\textwidth]{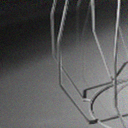}}
  \subfloat{\includegraphics[width=0.16\textwidth,height=0.13\textwidth]{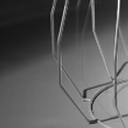}}
  \subfloat{\includegraphics[width=0.16\textwidth,height=0.13\textwidth]{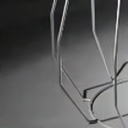}}
  \subfloat{\includegraphics[width=0.16\textwidth,height=0.13\textwidth]{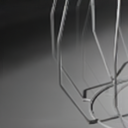}}
  \subfloat{\includegraphics[width=0.16\textwidth,height=0.13\textwidth]{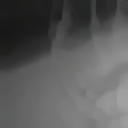}}
  \vspace{-.2cm}
  \caption{Burst Gaussian denoising with $\sigma=25$. Our method is able to effectively restore the images and retain fine details, as opposed to the rest of the methods that over-smooth high texture areas. Imprecise misalignment will cause methods to introduce visual artifacts such as those in the last row. Results best seen magnified on a computer screen.}
  \label{fig:burst_denoise}
\end{figure*}

{\small
\bibliographystyle{ieee}
\bibliography{egbib}

\begin{thebibliography}{10}\itemsep=-1pt

\bibitem{Aittala_2018_ECCV}
Miika Aittala and Fredo Durand.
\newblock Burst image deblurring using permutation invariant convolutional
  neural networks.
\newblock In {\em The European Conference on Computer Vision (ECCV)}, September
  2018.

\bibitem{beck2009fast}
Amir Beck and Marc Teboulle.
\newblock A fast iterative shrinkage-thresholding algorithm for linear inverse
  problems.
\newblock {\em SIAM journal on imaging sciences}, 2(1):183--202, 2009.

\bibitem{fista.2009}
Amir Beck and Marc Teboulle.
\newblock A fast {I}terative {S}hrinkage-{T}hresholding {A}lgorithm for
  {L}inear {I}nverse {P}roblems.
\newblock {\em SIAM Journal on Imaging Sciences}, 2(1):183--202, 2009.

\bibitem{buades2005non}
Antoni Buades, Bartomeu Coll, and J-M Morel.
\newblock A non-local algorithm for image denoising.
\newblock In {\em Computer Vision and Pattern Recognition, 2005. CVPR 2005.
  IEEE Computer Society Conference on}, volume~2, pages 60--65. IEEE, 2005.

\bibitem{jian.2009}
Jian-Feng Cai, Hui Ji, Chaoqiang Liu, and Zuowei Shen.
\newblock Blind motion deblurring using multiple images.
\newblock {\em Journal of Computational Physics}, 228(14):5057 -- 5071, 2009.

\bibitem{dabov2007image}
Kostadin Dabov, Alessandro Foi, Vladimir Katkovnik, and Karen Egiazarian.
\newblock Image denoising by sparse {3-D} transform-domain collaborative
  filtering.
\newblock {\em IEEE Transactions on image processing}, 16(8):2080--2095, 2007.

\bibitem{Danielyan_2009_BM3DCFA}
A. Danielyan, M. Vehvilainen, A. Foi, V. Katkovnik, and K. Egiazarian.
\newblock Cross-color bm3d filtering of noisy raw data.
\newblock In {\em 2009 International Workshop on Local and Non-Local
  Approximation in Image Processing}, pages 125--129, Aug 2009.

\bibitem{Evangelidis.2008}
G.~D. Evangelidis and E.~Z. Psarakis.
\newblock Parametric image alignment using enhanced correlation coefficient
  maximization.
\newblock {\em IEEE Transactions on Pattern Analysis and Machine Intelligence},
  30(10):1858--1865, Oct 2008.

\bibitem{farsiu.2004}
S. Farsiu, M.~D. Robinson, M. Elad, and P. Milanfar.
\newblock Fast and robust multiframe super resolution.
\newblock {\em IEEE Transactions on Image Processing}, 13(10):1327--1344, Oct
  2004.

\bibitem{Gharbi:2016:DJD:2980179.2982399}
Micha\"{e}l Gharbi, Gaurav Chaurasia, Sylvain Paris, and Fr{\'e}do Durand.
\newblock Deep {J}oint {D}emosaicking and {D}enoising.
\newblock {\em ACM Trans. Graph.}, 35(6):191:1--191:12, Nov. 2016.

\bibitem{godard2017deep}
Clement Godard, Kevin Matzen, and Matt Uyttendaele.
\newblock Deep burst denoising.
\newblock In {\em The European Conference on Computer Vision (ECCV)}, September
  2018.

\bibitem{hasinoff2016burst}
Samuel~W. Hasinoff, Dillon Sharlet, Ryan Geiss, Andrew Adams, Jonathan~T.
  Barron, Florian Kainz, Jiawen Chen, and Marc Levoy.
\newblock Burst photography for high dynamic range and low-light imaging on
  mobile cameras.
\newblock {\em ACM Transactions on Graphics (Proc. SIGGRAPH Asia)}, 35(6),
  2016.

\bibitem{He_2015_ICCV}
Kaiming He, Xiangyu Zhang, Shaoqing Ren, and Jian Sun.
\newblock Delving deep into rectifiers: Surpassing human-level performance on
  imagenet classification.
\newblock In {\em The IEEE International Conference on Computer Vision (ICCV)},
  December 2015.

\bibitem{Healey.1994}
G.~E. Healey and R. Kondepudy.
\newblock Radiometric ccd camera calibration and noise estimation.
\newblock {\em IEEE Transactions on Pattern Analysis and Machine Intelligence},
  16(3):267--276, March 1994.

\bibitem{heide2016proximal}
Felix Heide, Steven Diamond, Matthias Nie{\ss}ner, Jonathan Ragan-Kelley,
  Wolfgang Heidrich, and Gordon Wetzstein.
\newblock Proximal: Efficient image optimization using proximal algorithms.
\newblock {\em ACM Transactions on Graphics (TOG)}, 35(4):84, 2016.

\bibitem{heide2014flexisp}
Felix Heide, Markus Steinberger, Yun-Ta Tsai, Mushfiqur Rouf, Dawid Pajak,
  Dikpal Reddy, Orazio Gallo, Jing Liu, Wolfgang Heidrich, Karen Egiazarian,
  et~al.
\newblock Flexisp: A flexible camera image processing framework.
\newblock {\em ACM Transactions on Graphics (TOG)}, 33(6):231, 2014.

\bibitem{Henz.2018}
Bernardo Henz, Eduardo S.~L. Gastal, and Manuel~M. Oliveira.
\newblock Deep joint design of color filter arrays and demosaicing.
\newblock {\em Computer Graphics Forum}, 37(2):389--399, 2018.

\bibitem{Hirakawa.2005}
K. Hirakawa and T.~W. Parks.
\newblock Adaptive homogeneity-directed demosaicing algorithm.
\newblock {\em IEEE Transactions on Image Processing}, 14(3):360--369, March
  2005.

\bibitem{khasabi2014}
D. Khashabi, S. Nowozin, J. Jancsary, and A.~W. Fitzgibbon.
\newblock Joint {D}emosaicing and {D}enoising via {L}earned {N}onparametric
  random fields.
\newblock {\em IEEE Transactions on Image Processing}, 23(12):4968--4981, Dec
  2014.

\bibitem{Kokkinos_2018_ECCV}
Filippos Kokkinos and Stamatios Lefkimmiatis.
\newblock Deep image demosaicking using a cascade of convolutional residual
  denoising networks.
\newblock In {\em The European Conference on Computer Vision (ECCV)}, September
  2018.

\bibitem{kokkinos2018iterative}
F. {Kokkinos} and S. {Lefkimmiatis}.
\newblock Iterative joint image demosaicking and denoising using a residual
  denoising network.
\newblock {\em IEEE Transactions on Image Processing}, pages 1--1, 2019.

\bibitem{kostadin2007video}
D Kostadin, F Alessandro, and E KAREN.
\newblock Video denoising by sparse 3d transform-domain collaborative
  filtering.
\newblock In {\em European signal processing conference}, volume 149. Tampere,
  Finland, 2007.

\bibitem{Lefkimmiatis_2017_CVPR}
Stamatios Lefkimmiatis.
\newblock Non-local color image denoising with convolutional neural networks.
\newblock In {\em The IEEE Conference on Computer Vision and Pattern
  Recognition (CVPR)}, July 2017.

\bibitem{lefkimmiatis.2017}
Stamatios Lefkimmiatis.
\newblock Universal denoising networks : {A} {N}ovel {CNN} {A}rchitecture for
  {I}mage {D}enoising.
\newblock In {\em The IEEE Conference on Computer Vision and Pattern
  Recognition (CVPR)}, June 2018.

\bibitem{lefkimmiatis.2013}
S. Lefkimmiatis, P. Ward, and M. Unser.
\newblock Hessian {S}chatten-norm regularization for linear inverse problems.
\newblock {\em IEEE Transactions on Image processing}, 22(5):1873--1888, 2013.

\bibitem{levin.2011}
Anat Levin and Boaz Nadler.
\newblock Natural image denoising: Optimality and inherent bounds.
\newblock In {\em Computer Vision and Pattern Recognition (CVPR), 2011 IEEE
  Conference on}, pages 2833--2840. IEEE, 2011.

\bibitem{Lin2015}
Qihang Lin and Lin Xiao.
\newblock An adaptive accelerated proximal gradient method and its homotopy
  continuation for sparse optimization.
\newblock {\em Computational Optimization and Applications}, 60(3):633--674,
  Apr 2015.

\bibitem{ma2017waterloo}
Kede Ma, Zhengfang Duanmu, Qingbo Wu, Zhou Wang, Hongwei Yong, Hongliang Li,
  and Lei Zhang.
\newblock {Waterloo Exploration Database}: New challenges for image quality
  assessment models.
\newblock {\em IEEE Transactions on Image Processing}, 26(2):1004--1016, Feb.
  2017.

\bibitem{maggioni2011video}
Matteo Maggioni, Giacomo Boracchi, Alessandro Foi, and Karen Egiazarian.
\newblock Video denoising using separable 4d nonlocal spatiotemporal
  transforms.
\newblock In {\em Image Processing: Algorithms and Systems IX}, volume 7870,
  page 787003. International Society for Optics and Photonics, 2011.

\bibitem{menon.2009}
D. Menon and G. Calvagno.
\newblock Joint demosaicking and denoisingwith space-varying filters.
\newblock In {\em 2009 16th IEEE International Conference on Image Processing
  (ICIP)}, pages 477--480, Nov 2009.

\bibitem{mildenhall2018burst}
Ben Mildenhall, Jonathan~T Barron, Jiawen Chen, Dillon Sharlet, Ren Ng, and
  Robert Carroll.
\newblock Burst denoising with kernel prediction networks.
\newblock In {\em Proceedings of the IEEE Conference on Computer Vision and
  Pattern Recognition}, pages 2502--2510, 2018.

\bibitem{parikh2014proximal}
Neal Parikh, Stephen Boyd, et~al.
\newblock Proximal algorithms.
\newblock {\em Foundations and Trends{\textregistered} in Optimization},
  1(3):127--239, 2014.

\bibitem{robinson.2008}
A.~J. Robinson and Frank Fallside.
\newblock The {U}tility {D}riven {D}ynamic {E}rror {P}ropagation {N}etwork.
\newblock Technical Report CUED/F-INFENG/TR.1, Engineering Department,
  Cambridge University, Cambridge, UK, 1987.

\bibitem{romano2017little}
Yaniv Romano, Michael Elad, and Peyman Milanfar.
\newblock The little engine that could: Regularization by denoising ({RED}).
\newblock {\em SIAM Journal on Imaging Sciences}, 10(4):1804--1844, 2017.

\bibitem{roth.2005.fields}
Stefan Roth and Michael~J Black.
\newblock Fields of experts: A framework for learning image priors.
\newblock In {\em Computer Vision and Pattern Recognition, 2005. CVPR 2005.
  IEEE Computer Society Conference on}, volume~2, pages 860--867. IEEE, 2005.

\bibitem{rudin1992nonlinear}
Leonid~I Rudin, Stanley Osher, and Emad Fatemi.
\newblock Nonlinear total variation based noise removal algorithms.
\newblock {\em Physica D: nonlinear phenomena}, 60(1-4):259--268, 1992.

\bibitem{Tai-MemNet-2017}
Ying Tai, Jian Yang, Xiaoming Liu, and Chunyan Xu.
\newblock Memnet: A persistent memory network for image restoration.
\newblock In {\em Proceedings of International Conference on Computer Vision},
  2017.

\bibitem{Tikhonov:1963}
A.~N. Tikhonov.
\newblock Solution of incorrectly formulated problems and the regularization
  method.
\newblock {\em Soviet Math. Dokl.}, 4:1035--1038, 1963.

\bibitem{Venkatakrishnan.2013}
S.~V. Venkatakrishnan, C.~A. Bouman, and B. Wohlberg.
\newblock Plug-and-{P}lay priors for model based reconstruction.
\newblock In {\em 2013 IEEE Global Conference on Signal and Information
  Processing}, pages 945--948, Dec 2013.

\bibitem{li.2008}
Lei~Zhang Xin~Li, Bahadir~Gunturk.
\newblock Image demosaicing: a systematic survey.
\newblock volume 6822, pages 6822 -- 6822 -- 15, 2008.

\bibitem{DCNN}
K. Zhang, W. Zuo, Y. Chen, D. Meng, and L. Zhang.
\newblock Beyond a {G}aussian {D}enoiser: {R}esidual {L}earning of {D}eep {CNN}
  for {I}mage {D}enoising.
\newblock {\em IEEE Transactions on Image Processing}, 26(7):3142--3155, July
  2017.

\bibitem{zhang.2017}
K. Zhang, W. Zuo, S. Gu, and L. Zhang.
\newblock Learning {D}eep {CNN} {D}enoiser {P}rior for {I}mage {R}estoration.
\newblock In {\em 2017 IEEE Conference on Computer Vision and Pattern
  Recognition (CVPR)}, pages 2808--2817, July 2017.

\bibitem{liu2014burstdenoising}
Xiaoou Tang Matt~Uyttendaele Ziwei~Liu, Lu~Yuan and Sun Jian.
\newblock Fast burst images denoising.
\newblock {\em ACM Transactions on Graphics (TOG)}, 33(6), 2014.

\end{thebibliography}
}

\end{document}